\documentclass[12pt]{article}
\usepackage{booktabs}  
\usepackage{amsmath}
\usepackage{amsthm}
\usepackage{amssymb}
\usepackage[a4paper,margin=1in,footskip=0.25in]{geometry}
\usepackage{mathrsfs}
\usepackage{gensymb}
\usepackage{textcomp}  
\usepackage{xcolor}
\usepackage{lipsum}    
\usepackage{graphicx}
\usepackage{caption}
\usepackage{subcaption}
\usepackage{bm, bbm}
\RequirePackage{diagbox, adjustbox}
\usepackage{natbib}
\usepackage[colorlinks=true,linkcolor=black, citecolor=blue,  urlcolor=black]{hyperref}
\usepackage{multirow}
\usepackage{multicol}
\usepackage{pdflscape}
\usepackage{afterpage}
\usepackage{everypage}
\usepackage{float}
\usepackage{natbib}
\usepackage{url} 

\newcommand{\blind}{1}

\begin{document}

\def\spacingset#1{\renewcommand{\baselinestretch}%
{#1}\small\normalsize} \spacingset{1}


\if1\blind
{
  \title{\bf Decision Tree-Based Predictive Models for Academic Achievement Using College Students' Support Networks}
  \author{Anthony Frazier$^1$, Joethi Silva$^2$, Rachel Meilak$^3$, Indranil Sahoo$^4$\thanks{Corresponding author}, \\
  David Chan$^4$, and Michael Broda$^4$ \\ \\
    $^1$Weber State University\\ 
    $^2$George Mason University \\
    $^3$Loyola Marymount University\\
    $^4$Virginia Commonwealth University
    }
    \date{}
  \maketitle
} \fi

\if0\blind
{
  \bigskip
  \bigskip
  \bigskip
  \begin{center}
   {\LARGE\bf Decision Tree-Based Predictive Models for Academic Achievement Using College Students' Support Networks}
\end{center}
 \medskip
} \fi

\begin{abstract}
In this study, we examine a set of primary data collected from 484 students enrolled in a large public university in the Mid-Atlantic United States region during the early stages of the COVID-19 pandemic. The data, called \textit{Ties} data, included students' demographic and support network information. The support network data comprised of information that highlighted the type of support, (i.e. emotional or educational; routine or intense). Using this data set, models for predicting students' academic achievement, quantified by their self-reported GPA, were created using \textit{Chi-Square Automatic Interaction Detection} (CHAID), a decision tree algorithm, and \textit{cforest}, a random forest algorithm that uses conditional inference trees. We compare the methods' accuracy and variation in the set of important variables suggested by each algorithm. Each algorithm found different variables important for different student demographics with some overlap. For White students, different types of educational support were important in predicting academic achievement, while for non-White students, different types of emotional support were important in predicting academic achievement. The presence of differing types of routine support were important in predicting academic achievement for cisgender women, while differing types of intense support were important in predicting academic achievement for cisgender men.

\end{abstract}

\noindent%
\vfill

\newpage
\spacingset{1.45} 

\section{Introduction}
\label{sec:intro}
In modern day society, there is an increasing emphasis on education. Improving educational outcomes, therefore, has become a priority, both at the individual and institutional level. This goal has prompted many researchers to investigate factors that help predict one’s academic achievement,  which is often measured quantitatively by Grade Point Average (GPA). The factors that are critical to a student's success are likely varied and possibly numerous. College achievement has previously been associated with race and ethnicity of students \citep{fletcher2010race}. In addition to using demographic data, the impact of students' personal networks and perceived social support on their academic achievement has also been studied for various university and community college settings across the globe \citep{eggens2008influence,ChengWen2012Hifs,china2015relationship,zavatkay2015social,li2018social}.

Emerging in the 1930s, social network analysis began as a way for social psychologists to represent the way people are connected \citep{JohnScott1988TRSN}. Since then, social networks have been utilized by many disciplines. These networks have been associated with the mathematical field of Graph Theory due to their easy abstraction into a series of nodes (representing people) and edges (representing the interaction between two people).  More recently, they have become important in the field of education because of the ties found between one’s social network, such as family or friends, and one’s academic achievement. \cite{ChengWen2012Hifs} found that perceived social support from one’s family was an important predictor of the student’s GPA regardless of their gender. \cite{azizi2013study} found that family support only predicted the GPA of the female students studied, but that peer social support predicted the GPAs of both the male and female students studied. Furthermore, \cite{deberard2004predictors} found that perceived social support, measured as the total score from the Multidimensional Perceived Social Support Scale (MPSSS), was a significant predictor of academic achievement, highlighting the importance of considering one’s social support network when building a predictive model for academic excellence.

In this study, along with demographic information, information about students' support networks is included as factors to aid in understanding the underlying influences on academic achievement. The uniqueness of this study lies in the fact that it analyzes the effects of one's social support network structure, the strength of individual ties, and the range of support types (emotional and educational) on students' self-reported GPA. Previous research has specified different types of support \citep{ChengWen2012Hifs,azizi2013study}, but this study distinguishes between routine and intense emotional and educational support. Further differentiating this study is the nature of the statistical models utilized. Unlike the previous studies which primarily used multinomial logistic regression models when predicting GPA \citep{eggens2008influence}, this study pursues advanced data mining techniques. All analyses are done by creating decision trees using the Chi-Square Automatic Interaction Detection (CHAID) technique and Random Forests based on conditional inference trees (\textit{cforest}). 

Decision trees have become an increasingly popular way to create classification or regression models because of their ability to capture non-linear effects on the response variable \citep{mcardle2013contemporary}. They are also relatively easy to create and understand \citep{McCarthy2019}, making them a popular choice for many researchers. They are highly flexible because of their nonparametric nature and are known to work well with high-dimensional data \citep{BehrAndreas2020EPoU}. Decision tree algorithms use a form of an IF-THEN-ELSE statement to determine how best to split the data \citep{McCarthy2019}, forming nodes based on the variable that accounted for each split. The most popular decision tree algorithm is Classification and Regression Trees (CART), which creates binary trees using Gini’s impurity index as splitting criterion \citep{breiman1984classification}. While decision trees are easier to interpret, researchers have sought to improve the overall accuracy of these decision tree models by generating ``random forests''. 

Formally introduced by Ho and Breiman \citep{TinKamHo1995Rdf,breiman1999random}, a random forest algorithm functions by creating an ensemble of trees, predicting the response value of a given observation based on aggregated output of the collection of trees \citep{MohapatraNiva2020OotR}. Random forests get their name from utilizing bootstrapping to iteratively generate both training sets and sets of predictor variables considered in the model \citep{JamesGareth2013Aits}.   Recent applications of random forests include predicting flood susceptibility \citep{ChenWei2020Mfsu}, solar radiation \citep{BenaliL2019Srfu}, and stroke outcome \citep{Fernandez-LozanoCarlos2021Rfpo}. Random forests account for missing data in both training and testing sets \citep{breiman2004random}, and random forest predictions are resistant to outliers \citep{breiman2001random}. The \textit{cforest} algorithm \citep{StroblCarolin2007Birf} is an implementation of the random forest algorithm using conditional inference trees.

Research investigating student academic achievement including social network data through the use of decision tree models is currently limited. \cite{Al-BarrakMashaelA2016PSFG} used the J48 decision tree algorithm to predict final GPA, but this model used students' previous grades as predictors, not social network support data. Similarly, \cite{kovacic2010early} used CHAID and CART to predict a student's success, but generated their model using socio-demographic variables such as age, gender and ethnicity, as well as overall study environment. Our study uses many predictor variables from the student’s social support network, thereby distinguishing it from previous studies. 

Comparing multiple tree algorithms when creating a model has also become more prevalent in data mining research, as highlighted by recent literature. \cite{shirali2018predicting} compared CHAID and CART when creating models to predict the outcome of occupational accidents at a steel factory in Iran and found similar levels of accuracy between the two models. \cite{PITOMBO201716} compared CHAID and CART with gravity models to estimate intercity trip distribution, but found that the CHAID model had more accuracy. \cite{SUT201115534} compared CHAID, Exhausted CHAID (E-CHAID), CART, the Quick, Unbiased, Efficient Statistical Tree (QUEST) method, the Random Forest Regression and Classification (RFRC) method and the Boosted Tree Classifiers and Regression (BTCR) when creating models to predict mortality in head injuries. The study found that CART performed the worst while BTCR performed the best. In addition, studies such as \cite{venkatasubramaniam2017decision} and \cite{gomes_lemos_jelihovschi_2020} compared CART with the conditional tree model, providing mixed results in terms of advantage. Lastly, comparisons between single-tree models ($ID_{3}$, $C_{4.5}$) and random forests have previously been done in \cite{SathyadevanShiju2014CAoD}. 

Our study compares a different set of single-tree and random forest predictive models, CHAID and \textit{cforest}, making it unique. These algorithms were chosen based on the particulars of this study. CHAID produces non-binary trees, which were preferable due to the prevalence of non-binary predictor variables in our data set.  On the other hand, the \textit{cforest} model was chosen over the popular \textit{randomForest} algorithm \citep{breiman2001random} since it has been shown that \textit{randomForest} gives unreliable results when the number of classes in the predictor variable differs from that of the response variable, which is a characteristic of our data set \citep{StroblCarolin2007Birf}. 

\section{Ties Data: Data Description}%
\label{sec:data}

The data used in this study was collected through an online survey in Spring 2020 after institutions across the Unites States had shut down due to the COVID-19 pandemic. 484 undergraduate and graduate students from a mid-Atlantic US University participated in the survey. The survey included questions about demographics, mental health, support network structure, academic perseverance, and collaborative learning. Students were also asked to rate the impact of COVID-19 on their academic performance and mental health. The primary goal of the survey was to find relationships between students'  support network and its properties with their self-reported academic performance. 

The \textit{Ties data} consists of demographic variables such as race, ethnicity, age, gender, and major. Students' mental health is measured using the Depression, Anxiety, and Stress Scale (DASS-21), in which participants are asked to answer 21 questions on a 4-point Likert scale. Perceived social support is measured using the Multidimensional Scale of Perceived Social Support (MSPSS), in which participants are asked to answer 12 questions about perceived social support from family, friends/peer, and significant others on a 7-point Likert scale.

Data collection on students' support networks involved survey questions asking them to list between 5 to 10 people who gave them support in their academic year. For each nominee, students noted the role of the nominee (i.e. family, peer, significant other, etc.), amount of strain within the relationship (on a scale of 1-10), and closeness of relationship (on a scale of 1-3).  The study also classified the support into intense and routine emotional support, as well as intense or routine educational support. For each nominee, students also indicated the type(s) of support provided, if any. If a nominee provided a particular support, the helpfulness of the support is rated on a scale of 1-10.

\subsection{Data Collection}
The Ties data was collected during an eight-week period from April to June 2020. Participants were informed of the study via the university's events newsletter and daily digital news. Each participant was compensated for their participation with a \$20 gift card. All participants were current students at the university where this survey took place. Participants were required to complete a screening questionnaire following which they completed the survey using Research Electronic Data Capture, Version 4.14.4. 

The goal of the study is to utilize demographic and support network data to create predictive models for academic achievement. To accomplish this, observations with incomplete responses for social network or demographic information questions were discarded. Discounting such incomplete surveys left us with a sample size of 320 students. 

As part of data pre-processing, the categorical demographic and social network variables were first identified. The other predictor variables were converted to categorical variables. This was done since CHAID requires predictor variables to have at least two categories, although the categories do not have to be ordered \citep{KassG.V1980AETf}. We included three demographic variables, $race, major \text{ and } gi$ (gender identity), along with variables relating to the overall support network. In addition, we defined categorical variables which further describe each participant's support network. Denoted by $hm$ and $hf$, these variables indicate whether or not the students nominated their mother or father in their social network, respectively. We also used variables that counted the number of individuals with different closeness scores ($n4, n3, n2$). The final data set used in our analysis included a total of 39 predictor variables. Table \ref{table: p_demo} displayed below describes the demographic variables considered in our model. Tables \ref{table:binary_vars} and \ref{table:discrete_vars} describe the binary and multi-category support network predictor variables used in our models, respectively.

\begin{table}[H]
\caption{Demographic information of students who completed the online survey}
\label{table: p_demo}
\centering
 \begin{tabular}{||c  c  c  c||} 
 \hline
 Variable & Description & Variable's categories  & Percent \\ [0.5ex] 

 \hline\hline

gi & Student's gender &  & \\ \hline
   & & Cisgender Woman    & 77.19\% \\
   & & Cisgender man  & 18.23\% \\
   & & Gender Minority   & 4.69\% \\
\hline
race & Student's race &    &   \\ \hline
    & & Asian  & 18.13\% \\
    & & Black/Latinx/Other  & 39.06\% \\
    & & White  & 42.81\% \\ \hline
major & Student's major &  &      \\ \hline
    & & STEM & 51.56\% \\
    & & Non-STEM & 24.69\% \\ 
    & & No Major & 23.75\%\\[1ex] 
 \hline
 
 \end{tabular}
\end{table}

\begin{table}[H]
\caption{Description of binary support network variables used in our analysis}
\label{table:binary_vars}
\begin{centering}
\begin{tabular}{| m{1.5cm} | m{9.5cm} | m{2.5cm} |} 
 \hline
 Variable & Description  &    \% with property \\ [0.5ex] 
 \hline\hline

    $r_f$ & Presence of family   & 91.6\% \\
    \hline
    $r_p$ & Presence of peers   & 83.4\% \\
    \hline
    $r_{so}$ &  Presence of significant other   & 49.7\% \\
    \hline
    $r_o$ & Presence of other & 6.25\% \\
    \hline
    $hm$ & Presence of mother &  69.4\% \\
    \hline
    $hf$ & Presence of father &  48.1\% \\
    \hline
    $ed_{rf}$ & Presence of family routine educational support & 33.1\% \\
    \hline
    $ed_{if}$ & Presence of family intense educational support &  47.2\% \\
    \hline
    $em_{rf}$ & Presence of family routine emotional support &  60.6\% \\
    \hline
    $ed_{rp}$ & Presence of peer routine educational support & 30\%  \\
    \hline
    $em_{io}$ & Presence of other intense emotional support &  3.8\%  \\
    \hline
    $em_{ro}$ & Presence of other routine emotional support &  3.4\% \\
    \hline
    $ed_{io}$ & Presence of other intense educational support &  2.2\%  \\
    \hline
    $ed_{ro}$ & Presence of other routine educational support &  1.9\% \\
    \hline
    $em_{is}$ & Presence of significant other intense emotional support &  45.9\% \\
    \hline
    $em_{rs}$ & Presence of significant other routine emotional support &  43.1\% \\
    \hline
    $ed_{rs}$ & Presence of significant other routine educational support &  22.2\% \\
    \hline
    $em_{ip}$ & Presence of peer intense emotional support &  56.3\%   \\
    \hline
    $em_{rp}$ & Presence of peer routine emotional support & 56.3\%   \\
    \hline
    $ed_{ip}$ & Presence of peer intense educational support & 43.4\% \\ [1ex] 

 \hline
 
 \end{tabular}
 \end{centering}
 \end{table}

 \begin{table}[H]
 \caption{Description of multi-category support network variables used in our analysis}
 \label{table:discrete_vars}
 \begin{centering}
  \begin{tabular}{| m{1.5cm} | m{9cm}| m{1cm} |  m{1.3cm} |  m{0.8cm} |}
 \hline
 Variable & Description & Mean & Median & SD \\ [0.5ex] 
 \hline\hline

    $ed_{rc}$ & Total routine edu. supporters & 1.62 & 1 & 1.41 \\
    \hline
    $ed_{ic}$ & Total intense edu. supporters & 2.21 & 2 & 1.51 \\
    \hline
    $em_{rc}$ &  Total routine emo. supporters & 3.28 & 3 & 1.66 \\
    \hline
    $em_{ic}$ & Total intense emo. supporters & 4.03 & 4 & 1.62 \\
    \hline
    $n4$ & Total supporters with closeness rating 4 & 2.68 & 3 & 1.5 \\
    \hline
    $n3$ & Total supporters with closeness rating 3 & 1.39 & 1 & 1.09 \\
    \hline
    $n2$ & Total supporters with closeness rating 2 or less & 1.02 & 1 & 1.25 \\
    \hline
    $mx^1_{mc}$ & Max. emo. helpfulness from close supporters (3/4) & 8.88 & 10 & 2.01 \\
    \hline
    $mx^1_{dc}$ & Max. edu. helpfulness from close supporters (3/4) & 6.92 & 8 & 3.45 \\ 
    \hline
    $mx^2_m$ & 2nd largest emo. support from total network & 7.96 & 9 & 2.53 \\
    \hline
    $mx^1_m$ & Max. emo. support from total network & 9.04 & 10 & 1.80 \\
    \hline
    $mx^2_d$ & 2nd largest edu. support from total network  & 4.98 & 6 & 3.81 \\
    \hline
    $mx^1_d$ & Max. edu. support from total network & 7.61 & 8 & 2.98 \\

 \hline
 \end{tabular}
 \end{centering}
 \end{table}

 \subsection{Data Visualization}
 \label{subsec: data_vis}
We visualized the support network data using egocentric graphs. These graphs consisted of a central node, representing the students taking the survey and several other nodes representing nominees in the students' support network. The edges between the central node and the other nodes represented particular support(s) given to the student.  

\begin{figure}[H]
        \centering
        \begin{subfigure}[b]{0.495\textwidth}
            \centering
            \includegraphics[width=\textwidth]{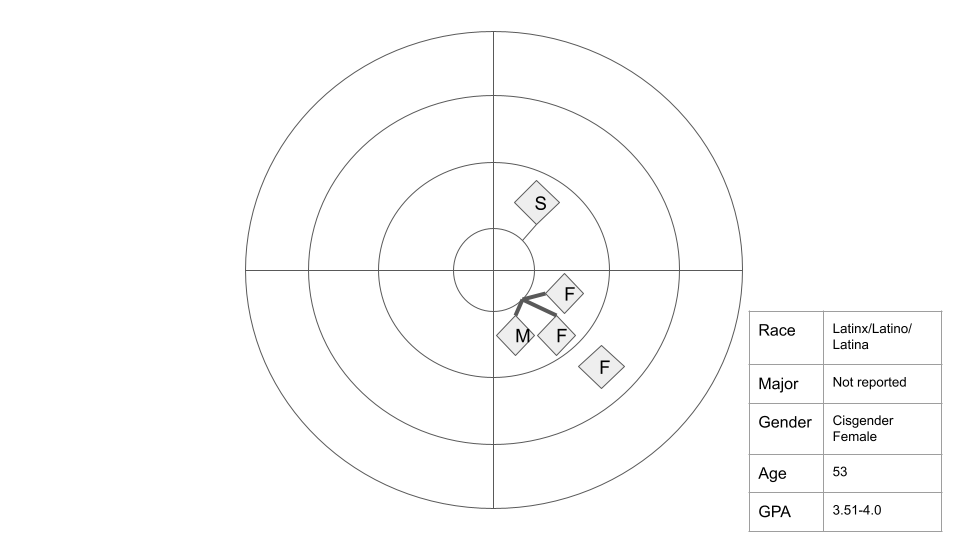}
            \caption[Network2]%
            {{\small Network 1}}    
        \end{subfigure}%
        \begin{subfigure}[b]{0.495\textwidth}  
            \centering 
            \includegraphics[width=\textwidth]{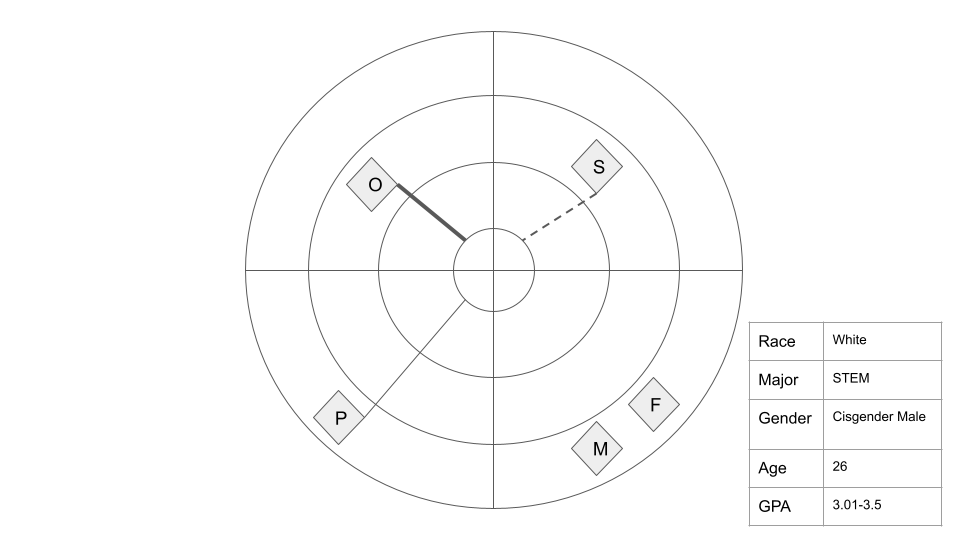}
            \caption[]%
            {{\small Network 2}}    
        \end{subfigure}%
        \vskip\baselineskip
        \begin{subfigure}[b]{0.495\textwidth}   
            \centering 
            \includegraphics[width=\textwidth]{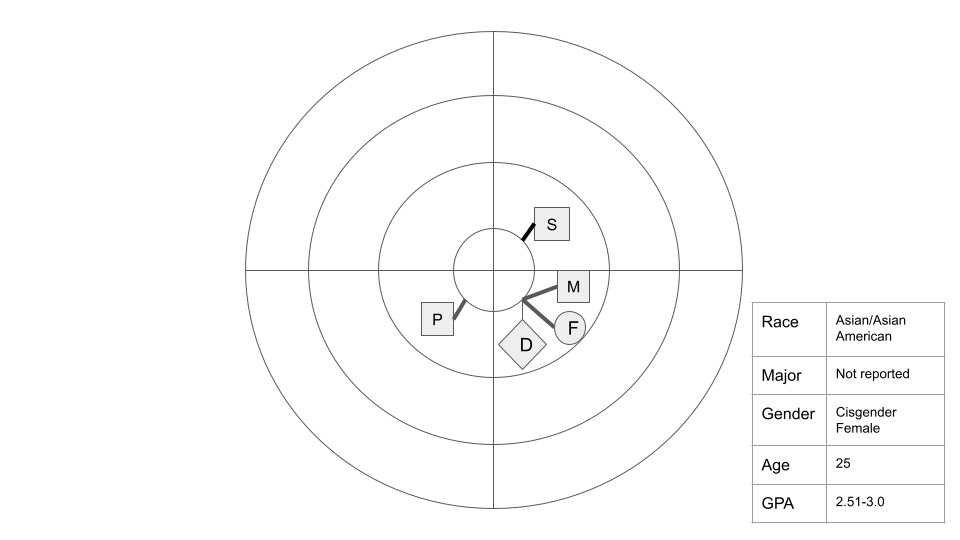}
            \caption[]%
            {{\small Network 3}}    
        \end{subfigure}%
        \begin{subfigure}[b]{0.495\textwidth}   
            \centering 
            \includegraphics[width=\textwidth]{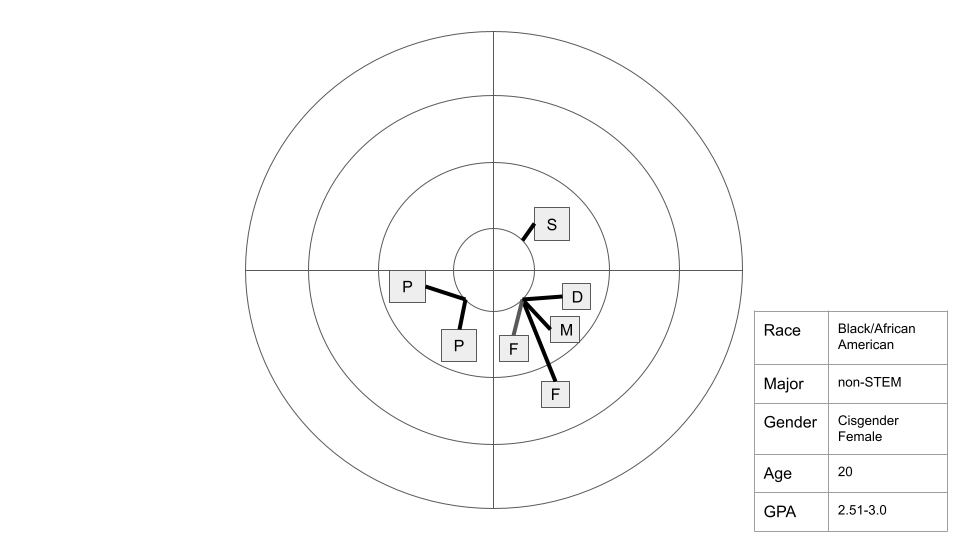}
            \caption[]%
            {{\small Network 4}}    
        \end{subfigure}%
        \caption[ Networks ]
        {\small Support networks for four students in the Ties data. The letters represent different individuals in the students' support networks (M = Mom, D = Dad, F = other family member, P = peer, S = significant other, O = other). The innermost circle in each network represents the student. Each tie is placed in one of the three rings depending on closeness of the individual nominated. The different shapes denote types of educational support provided (Diamond -- none, Circle -- routine, Triangle -- intense, Square -- both) and different line types denote types of emotional support provided (No line -- none, Dashed -- routine, Skinny solid -- intense, Thick solid -- both).} 
        \label{support_networks} 
\end{figure}

Individuals within a student's support network are divided into four categories: family (M = mom, D = dad, F = other family member), peer (P), significant other (S), and other (O).   The other category is a collection of people that do not fit in one of the first three categories.
Within each egocentric graph, the upper right quadrant is for ``significant others", the bottom left quadrant is for ``peers", the bottom right quadrant is for ``family",  and the upper left quadrant is for ``others". The shape of the nodes represent types of educational support provided (Diamond -- none, Circle -- routine, Triangle -- intense, Square -- both) while the different line types represent types of emotional support provided (No line -- none, Dashed -- routine, Skinny solid -- intense, Thick solid -- both). Finally, the innermost circle represents the participant, while the three outer rings represent varying levels of closeness ($ n4, n3$, and $n2 $) as we move from the center outwards. Four different support networks are shown in Figure \ref{support_networks}, highlighting the differences in networks for varying demographic and GPA groups.

As shown in Figure \ref{support_networks}, the student with Network 1 has many close supports, with a strong family support system. While no educational support is provided (all nodes are diamond shaped), there is strong emotional support provided (multiple nodes connected with thick solid lines). The student with Network 2 has no close supports and a more varied network, with support from all four quadrants. There is also no educational support provided, but there are varying amounts of emotional support (evident from the connections shown with different line types). Support Network 3 shows that the student has an extremely close network with many supporters aiding them in multiple ways highlighted by ties shown using squares and connected with thick solid lines. Finally, the student with Network 4 has a large and close network consisting exclusively of supporters that assist the student in every way.


\section{Methodology}%
\label{sec:methods}

We constructed two predictive models for predicting GPA: decision tree model using CHAID, and random forest model using \textit{cforest}.  We also conducted a model comparison between the CHAID and \textit{cforest} algorithms. This was accomplished by first comparing prediction accuracy of the CHAID and \textit{cforest} algorithms, then analyzing the differences in variables identified as important by the CHAID and \textit{cforest} algorithms.  All comparisons were done using the full data set and different subsets of data.  The data was divided into subsets by racial identities, gender identities and combinations of the two.  

\subsection{Chi-squared Automatic Interaction Detection (CHAID)}

The Chi-square Automatic Interaction Detector (CHAID) was developed by \cite{KassG.V1980AETf} and uses the chi-squared test of association as its splitting criterion, thus accounting for statistically significant associations between the response and predictor variables as the tree grows \citep{mcardle2013contemporary}. A Bonferroni correction on the p-value sets it apart from other methods \citep{mcardle2013contemporary}, as well as its ability to produce non-binary decision trees, which is one of the main reasons it has been chosen for this study. 

A CHAID tree is constructed by repeatedly splitting subsets of the data space into two or more child nodes \citep{michael1997data}. The best split at each node is determined by merging relevant pairs of categories of the predictor variables until there is no significant difference within the pair in terms of the response variable. The CHAID method automatically deals with interactions among predictor variables, and the final nodes correspond to different subgroups as defined by the sets of independent variables in the path leading to the respective nodes \citep{ho2004analysis}.

The CHAID algorithm for generating trees is especially useful in our study because our data contains many predictor variables with varying distributions, some of which are likely to interact with one another. Also, the CHAID algorithm relies on no underlying distributional assumptions and can produce models using many predictor variables with relative ease \citep{alma99130567583801101}.

\subsection{Random Forest using \textit{cforest} }
Random forests employ a large number of decision trees at training time and is widely used as a machine learning tool in classification and regression problems. A random forest comprises of decision trees made with randomly selected predictors. The predictions made by a random forest algorithm are obtained by summarizing over outputs from different trees in the forest. Random forests correct for the over-fitting issue that are prevalent in decision trees and hence predictions made by random forests often have higher precision. Random forests can also account for missing data and produce highly accurate classifiers for many data sets \citep{breiman2004random}.

The \textit{cforest} algorithm generates a random forest comprising of a collection of conditional inference trees. These trees utilize tests for independence between predictor variables and the response variable through the use of permutation test conditional distributions created by \cite{Strasser99onthe}. The conditional random forest algorithm works in three steps \citep{HothornTorsten2006URPA}. For each node in a conditional inference tree,
\begin{enumerate}
    \item Test the null hypothesis of independence for each predictor variable against the response variable. If the null hypothesis cannot be rejected, stop algorithm. Otherwise, select the predictor variable with the strongest association with the response variable
    \item Partition the chosen response variable into two disjoint sets (creating nodes).
    \item Recursively repeat steps 1 and 2. 
\end{enumerate}

The \textit{cforest} algorithm was chosen primarily due to its ability to work with predictor variables included in our data set. The predictor variables considered vary in their range of possible values. Some variables, such as $r_{f}$, are binary, others such as $race$ have several categories, which some, like $ed_{rc}$ and $mx^1_{mc}$, hold up to ten levels . While the CART-based \textit{randomForest} algorithm has selection bias towards variables with multiple categories or levels, this issue can be easily overcome using the \textit{cforest} algorithm \citep{StroblCarolin2007Birf}.

\subsection{Model Specifications}

The important parameters involved in the CHAID algorithm include a merging threshold ($\alpha_2$), denoting the significance level to determine which categories in a multi-category split should be merged, and a splitting threshold ($\alpha_4$), denoting the significance level to determine whether a node should be split. During our preliminary analysis, CHAID trees were generated using differing values of $\alpha_2$ and $\alpha_4$. It was found that for $(\alpha_2, \alpha_4) \in \lbrace 0.01, 0.03, 0.05 \rbrace \times \lbrace 0.01, 0.03, 0.05 \rbrace$, the CHAID trees tended to constitute a single node, that is, no predictor variables were found to be associated with the response variable at these significance levels. The alpha parameters are intended to be set to any conventionally accepted significance level for statistical tests \citep{mcardle2013contemporary}, so to encourage our tree to include splitting nodes, we set both $\alpha_2$ and $\alpha_4$ to be equal to $0.1$. 

Similarly, the important parameters characterizing the \textit{cforest} algorithm included the number of variables considered at each splitting node ($mtry$) and the number of trees generated for one forest model ($ntree$). 
In order to tune our model, a two-step optimization routine was implemented to determine optimal values of $ntree$ and $mtry$. In the first step, $ntree$ was optimized, keeping $mtry$ fixed at the square root of the number of predictor variables used in the model. The optimal $ntree$ was defined as that value of $ntree$ which maximized testing accuracy. The optimization was done over $ntree = \lbrace 1, 2, \ldots, 501\rbrace$. For the full data set, the optimal value for $ntree$ was found to be 175. In the second step, $mtry$ was optimized by maximizing a resampling based testing accuracy over $mtry = \lbrace 2, \ldots, 39\rbrace$, keeping $ntree$ fixed at 175. The second step was implemented using the \texttt{train} function in the \texttt{caret} package in \texttt{R} \citep{kuhn2008building}. While creating random forest models for subset data, $ntree$ was still considered to be 175. However, since the error rate is known to change with sample size \citep{janitza2018overestimation}, $mtry$ was optimized for every demographic subset to avoid possible underestimation of testing accuracy.

To more reliably compare the CHAID and \textit{cforest} algorithms, we set the common model parameters to be equal between both algorithms. These included the number of observations in splitted response where no further split is necessary ($minsplit$), and the minimum number of observations in each terminal node ($minbucket$). Since the sample sizes for our subset models were small, we chose relatively small values for $minsplit$ and $minbucket$ so as to keep them consistent through our different models for different subsets. In particular, $minsplit$ and $minbucket$ were chosen to be 10 and 2 respectively.

To better understand the overall accuracy of both the CHAID and \textit{cforest} models, training and testing set accuracy were calculated. For each demographic subset, both the \textit{cforest} and CHAID models were generated 500 times. New training and testing sets were used in each iteration. The average accuracy for each subset model was computed by taking the arithmetic mean of the accuracy from all iterations, 

Variables found important by the CHAID model were determined by selecting the variables present in the most representative CHAID tree. The important variables from the \textit{cforest} model were identified using the unbiased variable importance measure as described in \cite{StroblCarolin2007Birf}. The variable importance values for the \textit{cforest} model were normalized, so the variables found most important were variables at the upper end of the variable importance distribution. We compared the set of variables found important by the CHAID model to those found important by the \textit{cforest} model. Since the largest optimal CHAID tree among all our demographic subsets contained seven splitting nodes, we examined the seven most important variables determined by the \textit{cforest} model. 

All statistical analyses were conducted using the statistical software \texttt{R}. CHAID and random forest models were created using the \texttt{CHAID} \citep{KassG.V1980AETf} and \texttt{party} \citep{cforest1,StroblCarolin2007Birf,cforest3} packages in \texttt{R} respectively. All codes used to implement the CHAID and \textit{cforest} algorithms have been shared as supplemental material accompanying the online article.

\section{Results}
\label{sec:res}

The models constructed for predicting GPA are divided into four categories: overall, race, gender, and the combination of race and gender.  The overall results are based on the completed survey data without subsets.  The race models considered data broken down into three racial subsets that were based on the amount of available data:  White, Asian, and Black/Latinx/Other.  The gender models were created based on data divided into subsets by cisgender women and cisgender men.  There was not enough available data to include individuals outside of these two categories.  Finally, we  constructed models for cisgender women that were divided up into the racial categories of White, Asian, and Black/Latinx/Other.  This breakdown by race was not possible to do for cisgender men due to the small sample size.

Grade point average, GPA, was self-reported by students into categories.  The categories included below 2.0, 2.0 to 2.5, 2.5 to 3.0, 3.0 to 3.5, and 3.5 to 4.0. Based on the number of responses in each category of the data collected, we divided this response variable into two categories:  3.5 to 4.0, and 3.5 and below. Although there is some loss of information in terms of GPA, the focus on critical factors of successful students, those with GPAs above 3.5, can be compared with those students who are less successful, those with GPAs below 3.5, is still constructive.  These two distinct groups of students represent those who are very successful and those who have some room for improvement.

Training and testing accuracy were calculated for both models constructed using \textit{cforest} and CHAID.  These results are shown in the top panel of Table \ref{table:allacc}.

 The \textit{cforest} model had better testing and training accuracy than CHAID, which was expected. The \textit{cforest} algorithm is structured in a way such that multiple trees can relay information as they are being built, ensuring that trees become better as the algorithm progresses \citep{otte2013c}.  Generally \textit{cforest} is expected to do better than CHAID in regards to testing accuracy, but this was not always the case in our analysis. 

We also see that when we considered data from all the surveys with complete answers, the important common variables used between the models included major, race, gender identity, and whether the student nominated a mother in their support network. The important variables for each model are shown in Table \ref{table:impvar}.  For the \textit{cforest} algorithm, only the top seven variables of importance identified by the algorithm are listed.  For the CHAID algorithm, variables are listed in the typical order of levels used in the algorithm, i.e. the first variable or variables are the ones used at the first level, followed by those used at the next level. This, in a sense, gives them an importance ranking though there can be more than one variable in a single level. The different levels are demarcated by /'s in the tables listing the important variables. 

\subsection{Gender Subsets}
\label{subsec: gender}

The data was divided into three gender subsets:  cisgender women, cisgender men, and other.  Due to the lack on individuals that self-identified as a gender other than cisgender woman or man, no models were created for individual other than cisgender men and cisgender women. The second panel of Table \ref{table:allacc} shows the accuracies (in percentages) of both models for both subsets.  Here the \textit{cforest} model has a slightly better accuracy over the CHAID model. We also have a larger pool of data from cisgender women than cisgender men, which likely helps improve the accuracy of the models for cisgender women compared to those for cisgender men.

When looking at the gender subsets including other demographic data, the important variables as identified by each model are once again shown in Table \ref{table:impvar}. As seen in this table for cisgender women, major and race were once again recognized as important variables, similar to the models for overall data. The variables $n4$ and $ed_{rf}$ were also common between both models.  These reflect that the number of distant nominees, and the presence of routine educational support from a student's family are important variables for modeling high academic achievement for cisgender women.

For the cisgender men, the common variables included $n2$ and $em_{rc}$, which refer to the number of closest individuals in the support network and the number of routine emotional supporters in a student's network.  Since the data had a majority of cisgender women participating, it is not surprising that the important variables for the entire data were more aligned with those for the cisgender women and not the cisgender men.


\subsection{Race Subsets}
\label{subsec:race}

Next we divided the data based on the self-identified race of the student.  Due to sample sizes of the races within the data (see Table \ref{table: p_demo}), the self-reported GPA was modeled separately for the Asian, Black/Latinx/Other and the White students.  Black/Latinx/Other students are the combination of students with African/African American heritage with those with Latin or Hispanic heritage.  The third panel of Table \ref{table:allacc} shows the accuracy rates (in percentages) for subset models using both methods.

For all race subsets, CHAID had higher testing accuracy than \textit{cforest}, deviating from the trend seen thus far. \textit{cforest} performed the worst for Asian students, with a testing accuracy  of about $51\%$.  Both methods has the highest testing accuracies for the combined Black/Latinx/Other students.

The important variables for each race subset model is shown in the third panel of Table \ref{table:impvar}.  Examining this part of the table closely, we see that each racial group have different sets of important variables, and very few were in common between the two methods. For the Asian students, presence of routine emotional support from family members and intense emotional support from peers were identified as important variables in both models. For the Black/Latinx/Other students, the effectiveness of the second highest emotional support was identified as important in both models. For the White students, major, the total number of intense emotional supporters and presence of mother in their support networks were the common important variable in both models.


\subsection{Gender and Race Subsets}
\label{subsec: gender_race}

To get an in-depth understanding of factors influencing academic achievement among college students, combination subsets were created using race and gender. As previously noted, race subsets within cisgender women were the only ones assessed due to the small sample sizes of subsets for cisgender men or gender minority. Accuracy results (in percentages) for these subset models are shown in the last panel of Table \ref{table:allacc}.

\begin{table}
\caption{Training and testing accuracy results (in percentages) from fitting \textit{cforest} and CHAID models for the overall data set (Total), cisgender women and cisgender men (Gender), Asian, Black/Latinx/Other and White students (Race) and Asian, Black/Latinx/Other and White cisgender women students (Gender \& Race).}
    \begin{centering}
\begin{tabular}{| l | p{1.8cm} | p{1.5cm} | p{1.8cm} | p{1.5cm} |}
\hline
  & \multicolumn{2}{l|}{Training Accuracy (in \%)}                         & \multicolumn{2}{l|}{Testing Accuracy (in \%)}                         \\ \hline
& \multicolumn{1}{l|}{\textit{cforest}} & \multicolumn{1}{l|}{CHAID} & \multicolumn{1}{l|}{\textit{cforest}} & \multicolumn{1}{l|}{CHAID} \\ 
 \hline \hline
 \multicolumn{5}{|l|}{Total}\\
 \hline
 & 96.47  & 71.40  &  \textbf{61.2} & 59.44 \\ 
 \hline \hline
 \multicolumn{5}{|l|}{Gender}\\
 \hline
 Cisgender women &  95.77 & 74.88 & \textbf{64.10} & 62.61 \\ 
 \hline
 Cisgender men   & 96.46  & 70.54 & \textbf{62.03} & 54.37 \\ 
 \hline \hline 
 \multicolumn{5}{|l|}{Race}\\
 \hline
 Asian & 90.93 & 74.17 & 50.66 & \textbf{58.47} \\ 
 \hline
 Black/Latinx/Other & 96.49 & 73.50 & 58.38 & \textbf{61.17} \\ 
 \hline
 White & 96.86  & 69.08 & 55.02 & \textbf{56.05} \\ 
 \hline \hline
 \multicolumn{5}{|l|}{Gender \& Race}\\
 \hline
 Asian Cisgender women & 83.28 & 76.38 & \textbf{65.91} & 59.96 \\ 
 \hline
 Black/Latinx/Other Cisgender women & 95.13 & 73.73 & 58.03 & \textbf{63.01} \\ 
 \hline
 White Cisgender women & 95.06  & 75.54 & \textbf{63.21} & 61.72 \\ 
 \hline
\end{tabular} \label{table:allacc}
\end{centering}
\end{table}

Subsetting by a combination of race and gender changed the accuracy results to an extent. Neither model was uniquely better in regards to testing accuracy. For the Black/Latinx/Other cisgender women, CHAID performs better, but for the Asian and White cisgender women, \textit{cforest} performed better.  The testing accuracy for the Asian cisgender women subset increased over $15\%$ from the pure race subsets when using \textit{cforest}, highlighting the importance of looking at the combination subsets. 
    
The important variables for these smaller subset for each subset model are given in the last panel of Table \ref{table:impvar}. For Asian cisgender women students, nominating a father in their support network and presence of intense educational support from a significant other were deemed important in both models. For Black/Latinx/Other cisgender women, presence of a significant other in their support network and the number of routine emotional support were important in both models. Finally, for White cisgender women, major and the presence of routine emotional support were found important in both models.

\begin{table}
    \caption{Summary of important variables identified by the \textit{cforest} and CHAID models for the overall data set (Total), cisgender women and cisgender men (Gender), Asian, Black/Latinx/Other and White students (Race) and Asian, Black/Latinx/Other and White cisgender women students (Gender \& Race). The variables identified as important by both models have been highlighted in bold.}
    \begin{centering}
    \resizebox{\columnwidth}{!}{
    \begin{tabular}{| l|l|l |}
        \hline
         & \textit{cforest}  & CHAID \\
        \hline \hline
        \multicolumn{3}{|l|}{Total}\\
        \hline
        & \textbf{major}, \textbf{race}, $ed_{ic}$, \textbf{gi}, $\bm{hm}$, $n4$, $hf$  & \textbf{major}/$\bm{hm}$, $em_{rf}$, \textbf{race}/$r_{so}$, \textbf{gi}, $em_{rf}$  \\
        \hline \hline
        \multicolumn{3}{|l|}{Gender}\\
        \hline 
        Cisgender  women & \textbf{major}, \textbf{race}, $\bm{n4}$, 
        $hm$, $\bm{ed_{rf}}$, $em_{rf}$, $ed_{rc}$ & \textbf{major}/$ed_{rc}$, $ed_{rf}$, \textbf{race}/$\bm{n4}$, $\bm{ed_{rf}}$/$em_{rc}$ \\
        \hline
        Cisgender men & $n3$, $\bm{n2}$, $mx^2_{d}$, $mx^2_{m}$, $\bm{em_{rc}}$, $ed_{ic}$, $mx^1_{d}$ & $\bm{n2}$/$\bm{em_{rc}}$, $r_{p}$/$hm$ \\
        \hline \hline
        \multicolumn{3}{|l|}{Race}\\
        \hline 
        Asian & $mx^1_{m}$, $\bm{em_{rf}}$, $n4$, $mx^1_{mc}$, $ed_{rp}$, $\bm{em_{ip}}$ & $hm$/$em_{rc}$, $\bm{em_{rf}}$/$\bm{em_{ip}}$, $ed_{is}$\\
        \hline
        Black/Latinx/Other  & major, $em_{ic}$, $n4$, $ed_{ic}$, $hm$, $n3$, $\bm{mx^2_{m}}$ & $em_{rp}$/$hf$/$\bm{mx^2_{m}}$ \\
        \hline
        White  & \textbf{major}, $\bm{hm}$, $\bm{em_{ic}}$, $em_{rc}$, $hf$, gi, $n4$  & \textbf{major}, $ed_{ip}$, $em_{rf}$/$\bm{hm}$/$\bm{em_{ic}}$  \\
        \hline \hline
        \multicolumn{3}{|l|}{Gender \& Race}\\
        \hline 
        Asian Cisgender & $em_{rf}$, major, $ed_{rp}$, $\bm{ed_{is}}$, $n2$, $\bm{hf}$, $ed_{rf}$ & $\bm{ed_{is}}$/$\bm{hf}$ \\
        women & & \\
        \hline
        Black/Latinx/Other  & $n4$, major, $em_{ic}$, $\bm{r_{so}}$, $ed_{ic}$, $\bm{em_{rc}}$, $em_{io}$  & $\bm{r_{so}}$/$\bm{em_{rc}}$, $ed_{ip}$ \\
        Cisgender women & & \\
        \hline
        White Cisgender & \textbf{major}, $\bm{ed_{rf}}$, $hf$, $ed_{rp}$, $em_{rf}$, $mx^2_{d}$, $mx^1_{dc}$  & $\bm{ed_{rf}}$/\textbf{major}, $ed_{is}$  \\
        women & & \\
        \hline
    \end{tabular} \label{table:impvar}
    }
    \end{centering}
\end{table}


\section{Discussion}%
\label{sec:discuss}

In this paper, we presented models for predicting self-reported GPA using both the CHAID and \textit{cforest} algorithms. In order to effectively address the relevant observations within each subset, we break our conclusions down by race, gender, and a combination of the two.

\subsection{Overall data}

Examining the overall models, and in particular the common variables we find that gender identity, nominating a mother, and race were all important variables.
Gender identity as an important variable is partially supported by previous literature. \cite{FortinNicoleM2015LbbG} found that, from the 1980s to the 2000s, the majority of girls went from achieving a `B' to an `A' in high school courses, while the majority of boys continued to achieve a `B'. Moreover, \cite{sonnert2012women} found that, on average, women had a GPA 0.1 points higher than their male counterparts in the same department, where GPA was evaluated on a 4 point scale. As these studies highlight, there are differences between genders in regard to GPA, which is why $gi$ was deemed important when looking at the complete data set.

Nominating a mother in the social network was important in regard to academic achievement. \cite{https://doi.org/10.1111/j.1741-3737.2007.00375.x} found that, although a variety of pairings not involving both biological parents reduced GPA, time with a single father decreased GPA more than time with a single mother, highlighting the importance of a mother's presence. Our study did not ask students to specify whether or not their parents were together. We do know, however, that those who nominated their mother feel that their mothers play a role in their lives, which is why \cite{https://doi.org/10.1111/j.1741-3737.2007.00375.x}'s conclusions can be viewed as supportive of our results.

\begin{figure}
    \centering
    \includegraphics[scale=0.35]{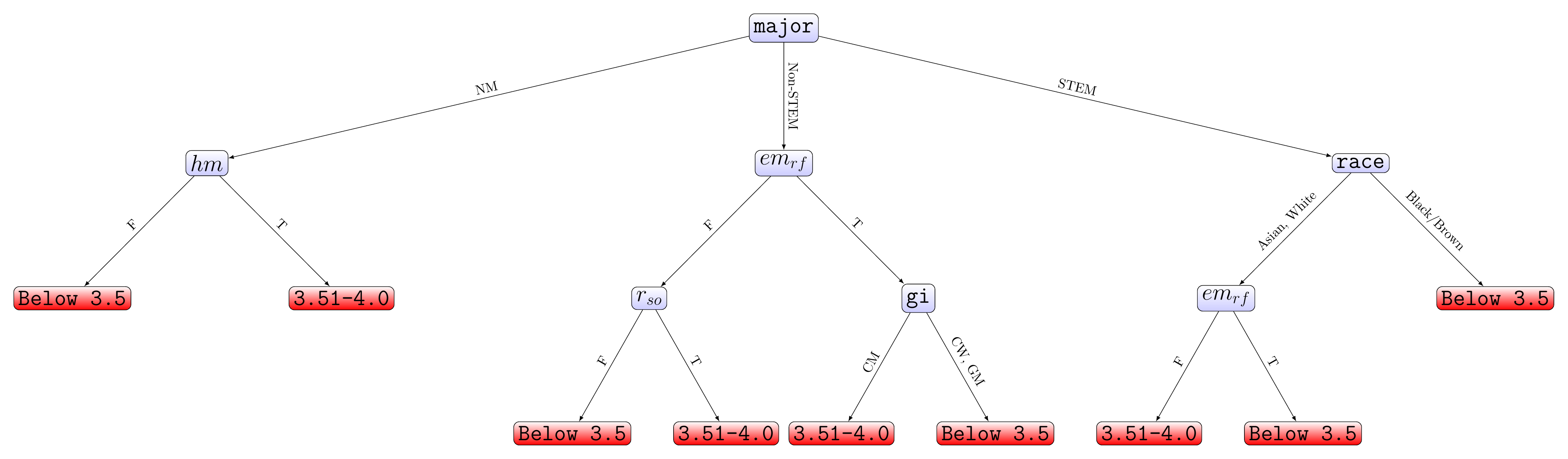}
    \caption{Optimal CHAID decision tree for self-reported GPA from obtained by using all predictor variables}
    \label{fig:nosubset_alldata}
\end{figure}

The race of the student was also found to be important when creating our models with the overall data. This is supported by prior research, since \cite{arcidiacono2012happens} found that Black students had a lower average GPA than non-Black students in the same field, regardless of both the type of field and the student's year (freshman, sophomore, junior, senior). Furthermore, \cite{doi:10.1080/10668926.2015.1112319} found that white men had higher GPAs then both Black and Latino men. These results support our findings, as shown in Figure \ref{fig:nosubset_alldata}. Students who were Black/Latinx/Other were automatically predicted to be in the lower GPA category, unlike their Asian and white counterparts, corroborating the results from the two studies mentioned.

\subsection{Variable Importance: Gender}

After analyzing the decision tree models, cisgender men and cisgender women only share two out of the eighteen variables that were found important in the decision trees. Thus, many variables found important in predicting GPA were unique to one gender subset, highlighting the different aspects of support important when assessing academic achievement for women and men. 

This is illustrated by the importance of the number of individuals at each closeness range in a social network. $n4$ represents the number of individuals with the closest rating of 4, which was found by both CHAID and \textit{cforest} to be important for cisgender women. $n4$ was not, however, important in either algorithm for cisgender men. Similarly, $n3$ and $n2$, or a closeness rating of 3 or less were important for cisgender men, but not for cisgender women. Therefore, while it is apparent that the overall closeness of the individual is important in regards to predicting academic achievement, regardless of gender, the specific level of closeness deemed important varies between cisgender women and cisgender men. 

The absence of peer support was common across subsets. $r_{p}$, or whether or not there was a peer in the social network, was found important by CHAID for the cisgender men. Neither algorithm, however, found any specific type of support from peers to be important for the cisgender men or cisgender women in predicting academic achievement. This finding is supported by \cite{doi:10.2190/A465-356M-7652-783R}, who found that peer support was unrelated to GPA. 

Furthermore, both CHAID and \textit{cforest} found the student's major was important for cisgender women, but not for cisgender men. Women in STEM must contend with sexism and biases that their male-counterparts do not, which could explain why major was only found important for cisgender women. For example, \cite{KuchynkaSophieL2018HaBS} found that women who experienced protective paternalism or hostile sexism frequently had lower STEM GPAs, specifically for women who were weakly identified with STEM.

Although cisgender women and cisgender men did share a few variables in common, variables associated with routine support were found important for cisgender women more often, while variables associated with intense support were more often found important for cisgender men. $ed_{ic}$, or the number of people in the social network who provided intense educational support, was important in \textit{cforest} for cisgender men.  
This variable was not found important by either algorithm for cisgender women, nor was any other variable associated with intense educational support. Furthermore, the only variable associated with routine support found important for cisgender men was $em_{rc}$, or the number of people in the social network who provided routine emotional support. \textit{cforest} and CHAID found this important for  cisgender men. However, $em_{rc}$ was also important in CHAID for cisgender women, as well as three other variables associated with routine support: $ed_{rc}$ (both CHAID and \textit{cforest}), $ed_{rf}$ (both CHAID and \textit{cforest}), and $em_{rf}$ (\textit{cforest}). Thus, although variables associated with both intense and routine support were found important for cisgender women, there was a clear trend of intense support being more important for cisgender men than routine support. 

\subsection{Variable Importance, Race}

Although the source varied, routine emotional support was important when predicting GPA for all race subsets. CHAID found that $em_{rp}$, the presence of routine emotional support from peers, was important for Black/Latinx/Other (\textit{cforest}) students. Both CHAID and \textit{cforest} found that $em_{rf}$, the presence of routine emotional support from family members, was important for Asian students.
The importance of emotional support as a predictor of GPA coincides with recent literature. \cite{li2020} found that emotional support from parents positively predicted college student's $4$-year GPA, as well as moderating the effect of race-related barriers on GPA. Furthermore, when looking at high school students, \cite{kashy2018predicting} found that emotional support provided by homeroom teachers was positively correlated with a student's GPA at the individual-level.

It should also be noted that intense educational support and routine emotional support were important regardless of race. CHAID found that $ed_{ip}$, the presence of intense educational support from a peer, was important for White students. For Black/Latinx/Other students, \textit{cforest} found $ed_{ic}$ important, the number of people in the social network who provided intense educational support. CHAID found $ed_{is}$, the presence of intense educational support from a significant other, important for Asian students. Regarding routine emotional support, $em_{rc}$ was significant for both Asian (CHAID) and the White students (\textit{cforest}), while $em_{rp}$ was significant for Black/Latinx/Other students (CHAID). These variables represent the number of people who provided routine emotional support and the presence of routine emotional support from a peer in the network, respectively, highlighting the importance of routine emotional support across race subsets.

\begin{figure}
    \centering
    \includegraphics[scale=0.6]{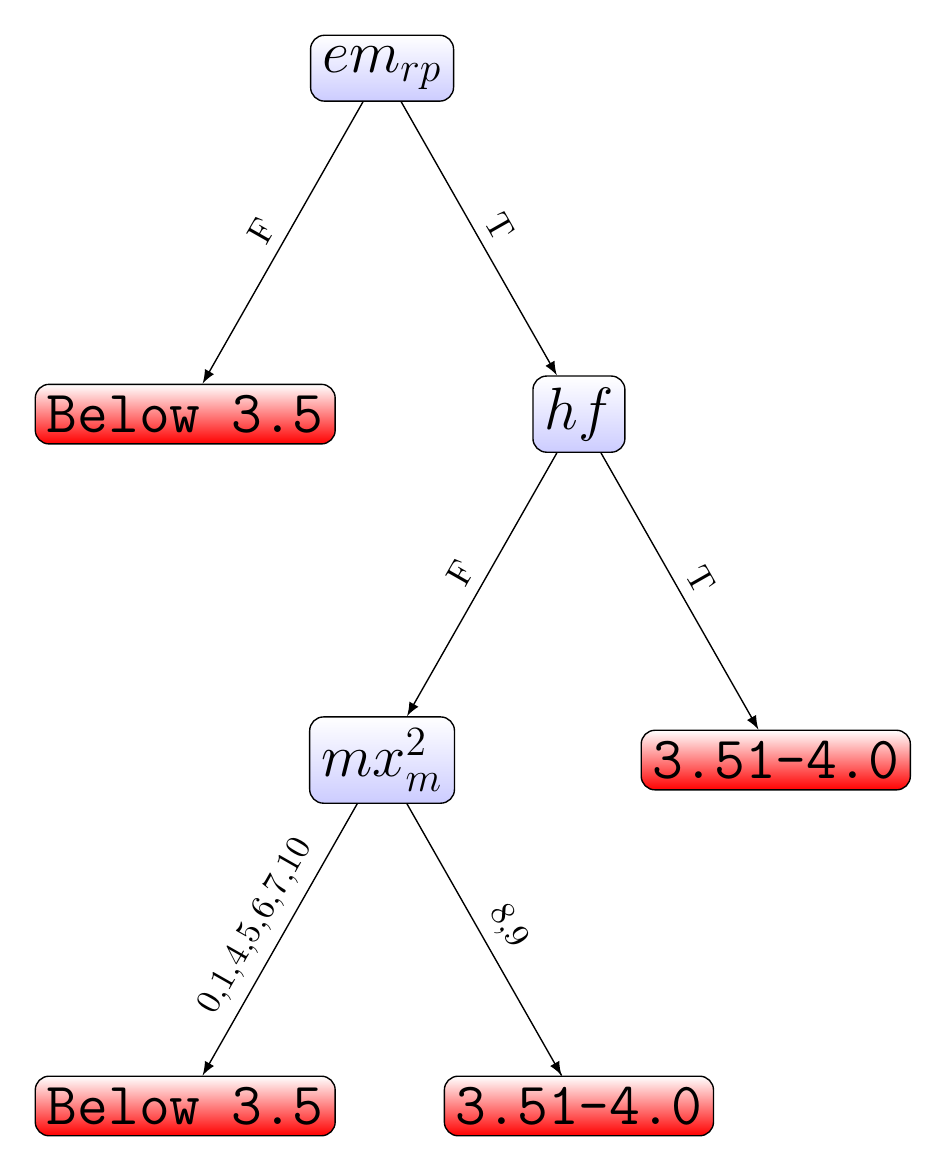}
    \caption{Optimal CHAID decision tree for self-reported GPA for the Black/Latinx/Others subset}
    \label{fig:blbr_alldata}
\end{figure}

The level of emotional support provided, however, was only important for minority racial groups. $mx^2_m$, the level of the second most important emotional support, was important for Black/Latinx/Other students in CHAID, while $mx^1_m$, the level of the first most important emotional support, was important for Asian students in \textit{cforest}. As shown in Figure \ref{fig:blbr_alldata}, having a higher level of the second most emotional support generally predicted higher academic achievement. This is partially supported by existing literature. \cite{kashy2018predicting} found that the perceived emotional support from homeroom teachers was positively related to GPA. Furthermore, \cite{li2020} found that emotional support from parents was positively correlated with GPA and helped to moderate race-related barriers to GPA. Emotional support, regardless of the source, appears to be positively related to GPA, supporting our findings. \cite{li2020}'s results specifically, however, could explain why having a high level of emotional support was important for minority racial groups who have historically faced more barriers to their education, but further research would have to be conducted to solidify this claim.  

\subsection{Variable Importance, Race + Gender}

The presence of different types of peer support was important when predicting GPA for various demographic subsets. First, $ed_{rp}$, the presence of routine educational support from peers was found important for the White cisgender women model (\textit{cforest}) and the Asian cisgender women model (\textit{cforest}). Furthermore, CHAID found that $ed_{ip}$, the presence of intense educational support from peers, was important for the Black/Latinx/Other model.

With the exception of white cisgender women, each model that found varying types of peer support  important in predicting GPA were from demographic-minority subsets. Existing literature supports our findings. First, a study on Asian and Hispanic sophomore college students found the lack of needed peer support had a negative correlation with future GPA, college adjustment and college committment \citep{DennisJessicaM2005TRoM}. Next, in the context of medical education, \cite{WebberSarah2021FaPS} found minority students tended to score statistically significantly lower than White students in perceived peer support, and as a result, had lower competency scores. Lastly, \cite{TuckerKathryn2020FHUS} determined academic-focused peer support programs improved the educational outcomes of students, with positive effects concentrated in racial minority groups. The research conducted by \cite{TuckerKathryn2020FHUS} concretely supports our findings with $ed_{rp}$ and $ed_{ip}$, and thus our results with these variables supplement current knowledge.

The presence of family support was also present across different subsets. The two variables found important relating to the presence of different supports from family were $ed_{rf}$ (presence of routine educational support from family) and $em_{rf}$ (presence of routine emotional support from family). Both variables were found important specifically for the white cisgender women and Asian cisgender women models. Existing literature partially supports these findings. First, \cite{ChengWen2012Hifs} found social support from family members was more important to women's academic achievement compared to men, explaining why $ed_{rf}$ and $em_{rf}$ were variables found important in specifically the female subsets of racial subsets as opposed to the racial subsets including men and women. Next, previous research suggests African-American students \citep{BrooksJadaE2015TIoF} and Hispanic students \citep{FiebigJenniferNepper2010HCCS} saw increases in GPA and overall academic achievement with increases and improvements of support from family. For our data, we did not find family support to be significant for these populations, so our results specifically for Black/Latinx/Other cisgender women are in opposition to the current research body.

\section{Concluding Remarks}
\label{sec: end}

In this study, we presented decision-tree models using CHAID and \textit{cforest} to determine academic achievement. This research included support network data to aid in determining academic achievement beyond basic demographic information. The addition by breaking up the data into subsets based on select demographics \textemdash race, gender, and a combination of the two \textemdash improved the accuracy of our models, as well as provided important information regarding the similarities and differences in variable importance between specific subsets. 

The type and source of support differed by demographic subset. For white students, different types of educational support were more important in predicting academic achievement than for non-White students, where emotional support played a larger role. Similarly, intense forms of support were more important for cisgender men than routine forms of support in predicting academic achievement, especially compared to cisgender women, where variables associated with routine support was more apparent. However, no concrete causal relationship can be assumed between the variables deemed important by the algorithms and academic achievement, especially due to the limitations in this particular study.

A limitation of our study comes from the limited scope of inference for our results. First, only one American university was used to survey participants, limiting responses to one region of the country. Previous research reports differences in students' social networks between rural and urban communities \citep{FungKaYi2015Ndae}, supporting the claim that our scope of inference is limited.

Second, due to the small sample size, the robustness of the variable importance results is limited. A small sample size allows for random chance/sampling error to play a larger role in explaining variability in our data. As a result of small sample size, our study is unable to make assertions about important social network variables in regard to academic achievement for demographic subgroups such as gender minority, cisgender Asian men, cisgender white men, and cisgender Black/Latinx/Other men. In addition, to effectively analyze students of color, we needed to group Black/African American, Latina/Latino/Latinx, and students with a race categorized as other into one demographic group: Black/Latinx/Other students. There are likely differences in social network structures and important variables between these student communities, and we are unable to provide inferences for them separately.

Additionally, a small sample size diminished the testing accuracy of both algorithms. 
The data collected for this analysis originated from an online survey. Some information about a student's social network would be better collected through qualitative questioning.  With an in-person survey, investigatory questions regarding the severity/type of strain, closeness and quality could be asked, providing clearer understanding of the nature of a student's support ties. In addition to question limitations, the online survey only required a minimum of five individuals in the support network section. As a result, many students only provided five individuals, though the survey permitted up to ten. Limiting the number of supporters analyzed prevented a more complete analysis of student's support networks to be conducted, which may have provided additional information that could help predict students' academic achievement.

Due to the limitations of this particular study, we propose further research on the properties of the CHAID and \textit{cforest} algorithms, as well as the differences in important support structure variables across different race and demographic groups. Analysis on the conditions under which \textit{cforest} outperforms CHAID and under which CHAID outperforms \textit{cforest} also deserves more attention. While we observed this variability within our data, we currently lack an explanation of why these differences exist between the distribution of variables for the racial subsets and gender subsets of our model. Additional research should examine the effects of altering sample size, distribution of predictor variables, as well as the number of levels for the response variable when determining under which context each model shows higher testing accuracy compared to the other.

The support network variables found important across our models apply to a limited population. We also propose, therefore, that further research should investigate what support variables are deemed important by these algorithms for samples collected across the United States and other countries, as well as more racially diverse samples. Aggregating results from a variety of samples will allow for stronger claims to be made about specific support network variables found important for specific demographic subsets. Once confidently claimed through statistical tests, important support network variables for specific demographic subsets should be investigated and verified through research in psychology and social science.

\section*{Supplementary Material}

\noindent Supplemental material linked to the online version of the paper includes \texttt{R} codes implementing the CHAID and \textit{cforest} algorithms and an example dataset used to demonstrate the codes.

\section*{Acknowledgement}
The work on this paper was supported by the National Science Foundation grant number DMS-1950015 and the National Security Agency grant number H98230-20-1-0011. Additionally, Chan was supported by a Virginia Commonwealth University College of Humanities and Sciences Catalyst grant.

\bibliographystyle{agsm}

\bibliography{GPA_arxiv}
\end{document}